\documentclass{article}
\usepackage{amsmath}
\usepackage{graphicx}

\usepackage[preprint]{neurips_2025}

\usepackage[utf8]{inputenc} 
\usepackage[T1]{fontenc}    
\usepackage{hyperref}       
\usepackage{url}            
\usepackage{booktabs}       
\usepackage{amsfonts}       
\usepackage{nicefrac}       
\usepackage{microtype}      
\usepackage{xcolor}         

\title{Network Inversion for \\Uncertainty-Aware Out-of-Distribution Detection}

\author{%
  Pirzada Suhail\\
  IIT Bombay\\
  \texttt{psuhail@iitb.ac.in} \\
  \And
  Rehna Afroz\\
  IIT Bombay\\
  \texttt{rafroz@iitb.ac.in} \\
  \And
  Gouranga Bala\\
  IIT Bombay\\
  \texttt{gbala@iitb.ac.in} \\
  \And
  Amit Sethi \\
  IIT Bombay\\
  \texttt{asethi@iitb.ac.in} \\
}

\begin{document}

\maketitle

\begin{abstract}

Out-of-distribution (OOD) detection and uncertainty estimation (UE) are critical components for building safe machine learning systems, especially in real-world scenarios where unexpected inputs are inevitable. However the two problems have, until recently, separately been addressed. In this work, we propose a novel framework that combines network inversion with classifier training to simultaneously address both OOD detection and uncertainty estimation. For a standard n-class classification task, we extend the classifier to an (n+1)-class model by introducing a "garbage" class, initially populated with random gaussian noise to represent outlier inputs. After each training epoch, we use network inversion to reconstruct input images corresponding to all output classes that initially appear as noisy and incoherent and are therefore excluded to the garbage class for retraining the classifier. This cycle of training, inversion, and exclusion continues iteratively till the inverted samples begin to resemble the in-distribution data more closely, with a significant drop in the uncertainty, suggesting that the classifier has learned to carve out meaningful decision boundaries while sanitising the class manifolds by pushing OOD content into the garbage class. During inference, this training scheme enables the model to effectively detect and reject OOD samples by classifying them into the garbage class. Furthermore, the confidence scores associated with each prediction can be used to estimate uncertainty for both in-distribution and OOD inputs. Our approach is scalable, interpretable, and does not require access to external OOD datasets or post-hoc calibration techniques while providing a unified solution to the dual challenges of OOD detection and uncertainty estimation.

\end{abstract}

\section{Introduction}
The increasing deployment of machine learning models in high-stakes, real-world applications—such as autonomous driving, medical diagnosis, and financial decision-making—has underscored the importance of model reliability and robustness. A key limitation of modern neural networks is their tendency to produce overconfident predictions \cite{suhail2025networkinversiongeneratingconfidently} even on inputs that lie far outside the training distribution. This makes it crucial to develop models capable of both out-of-distribution (OOD) detection—the ability to identify inputs that fall outside the training distribution—and uncertainty estimation (UE)—the ability to quantify confidence in predictions to ensure safe decision-making under distributional shift.

Both capabilities are vital for trustworthiness in deployment scenarios where the data encountered during inference may deviate from the training distribution in subtle or unexpected ways. Although these two problems are inherently linked, most existing approaches treat them separately, often relying on post-hoc calibration techniques or auxiliary OOD datasets, which may not always be available.

In this work, we propose a novel framework that leverages network inversion\cite{suhail2024networkcnn}, not only to detect OOD inputs but also to estimate prediction uncertainty, unifying the two objectives in a single training procedure. By extending a standard (n+1)-class model with an auxiliary garbage class, and iteratively refining the model using inverted reconstructions, we encourage the network to carve out clean decision boundaries while isolating ambiguous or anomalous regions. Unlike prior approaches, our method requires no external OOD datasets or post-hoc calibration, offering a simple and interpretable solution to ensure robustness in classification under distributional shift.

\section{Prior Work}

Inversion attempts to reconstruct inputs that elicit desired outputs or internal activations of a neural network. Early studies on multilayer perceptrons applied gradient-based inversion to visualize decision rules, but these often yielded noisy or adversarial-like images~\cite{KINDERMANN1990277,784232,SAAD200778}. Evolutionary search and constrained optimization were explored as alternatives~\cite{Wong2017NeuralNI}. Later work introduced prior-based regularization, including smoothness constraints and pretrained generative models, to improve realism and interpretability of reconstructions~\cite{mahendran2015understanding,yosinski2015understanding,mordvintsev2015inceptionism,nguyen2016synthesizing,nguyen2017plug}. The connection to adversarial examples has been emphasized, as unconstrained inversion can converge to adversarial artifacts~\cite{szegedy2013intriguing,goodfellow2014explaining}. In contrast, adversarially robust classifiers tend to produce more human-aligned features~\cite{tsipras2018robustness,engstrom2019adversarial}, enabling more interpretable reconstructions~\cite{santurkar2019image}. Recent advances include learning surrogate loss landscapes to stabilize inversion~\cite{liu2022landscapelearningneuralnetwork}, and generative methods that conditionally reconstruct inputs likely to produce a given output~\cite{suhail2024networkcnn}. Alternative formulations recast inversion into logical reasoning frameworks, encoding classifiers into CNF constraints for deterministic reconstruction~\cite{suhail2024network}.

Uncertainty quantification (UQ) has emerged as a cornerstone of reliable AI systems, particularly in domains where overconfident false predictions can lead to critical failures. Post-hoc methods are attractive because they can be retrofitted to pretrained deterministic classifiers without requiring retraining. Monte Carlo Dropout (MC Dropout) \cite{gal2016dropout} introduces stochasticity during inference to approximate Bayesian model averaging, while temperature scaling \cite{guo2017calibration} improves calibration with a single scalar parameter applied to logits. More recently, auxiliary prediction heads or meta-models have been explored. Evidential Deep Learning \cite{sensoy2018evidential} reformulates classification into the prediction of Dirichlet parameters, providing both predictive means and uncertainty. Direct Epistemic Uncertainty Prediction (DEUP) \cite{jain2022deup} learns a secondary model to estimate generalization error from data embeddings. Later, \cite{shen2023posthoc} proposes evidential meta-models that generate Dirichlet distributions from classifier logits. 

Bayesian neural networks (BNNs) \cite{neal1996bayesian} \cite{blundell2015weight} and related variational inference techniques  offer a more principled alternative by maintaining posterior distributions over network weights. Ensemble learning remains one of the most empirically effective strategies for UQ. Deep Ensembles \cite{lakshminarayanan2017simple} aggregate predictions from independently trained networks and consistently achieve strong calibration and robustness under distributional shift. Domain-specific strategies include test-time augmentation to approximate prediction variance, uncertainty-aware segmentation masks to enhance interpretability \cite{jungo2020uncertainty}, and Bayesian approximations adapted to volumetric imaging \cite{kwon2020uncertainty}. \cite{shen2023posthoc} proposed evidential meta-models trained on classifier embeddings to predict Dirichlet distributions, enabling decomposition into epistemic and aleatoric uncertainty. \cite{jain2022deup} generalized this idea with DEUP to out-of-distribution and low-data regimes, while  \cite{bala2025baymed} introduced BAY-MED, a Dirichlet meta-model for breast cancer classification that demonstrates robustness to OOD samples.

Recent work in \cite{ansari2022autoinverse} proposed Autoinverse, a framework for neural network inversion that prioritizes solutions near reliable training samples, using embedded regularization and predictive uncertainty minimization to improve robustness. Later \cite{fanlu2023scood} introduced a semantically coherent OOD detection (SCOOD) approach by combining uncertainty-aware optimal transport with dynamic cost modeling and inter-cluster enhancements. While \cite{chen2024uncertainty} developed a Gaussian process-based model that operates solely on in-distribution data. Similarly, \cite{charpentier2020postnet} presents PostNet, which employs normalizing flows to model posterior distributions over predicted probabilities, allowing reliable uncertainty estimation and effective OOD discrimination—even without OOD supervision.

\begin{figure*}[!t]
\centering
\includegraphics[width=\textwidth]{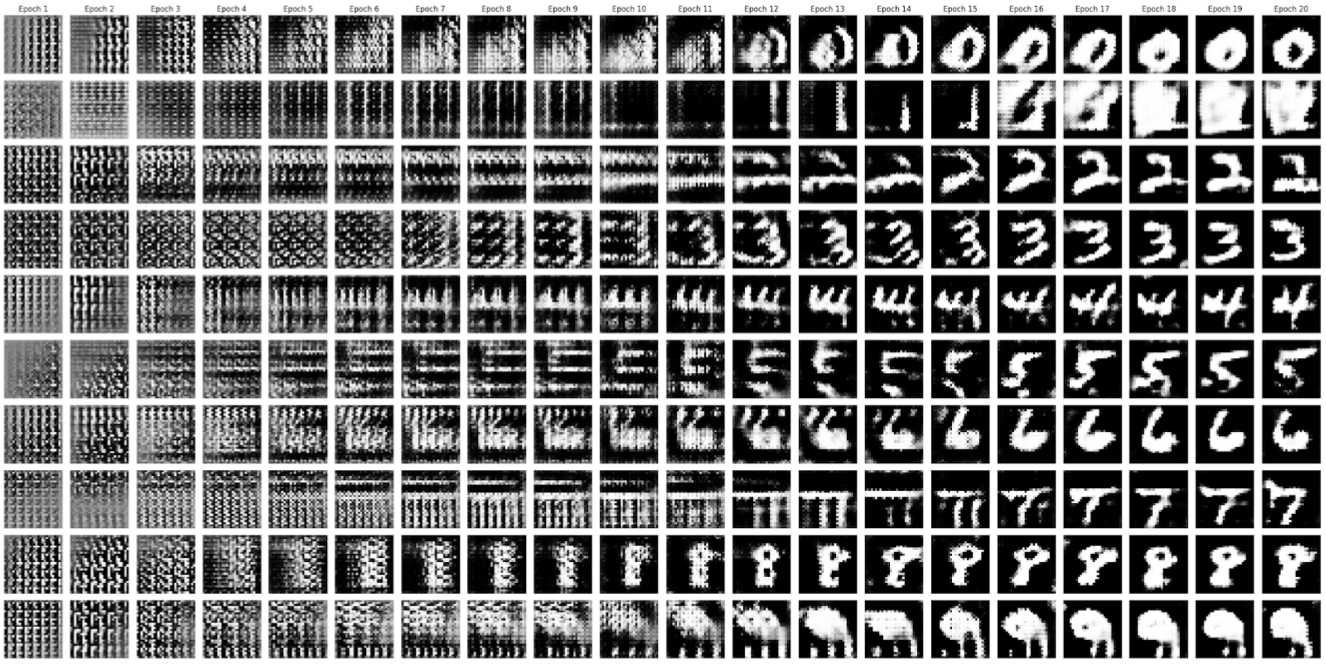}
\caption{Inverted Samples across epochs for different classes, beginning to resemble the training data as OODs are excluded into garbage class.}
\label{7}
\end{figure*}

\section{Methodology}
Our unified training approach integrates out-of-distribution (OOD) detection and uncertainty estimation (UE) into a single framework using network inversion and an auxilary garbage class. For an n-class classification task, we extend the classifier to an (n+1)-class model by introducing an additional "garbage" class designed to absorb anomalous inputs. This garbage class is initially populated with random Gaussian noise, representing OOD samples.

Between successive training epochs, we perform network inversion as in \cite{suhail2024networkcnn} to reconstruct samples from the input space of the classifier for all output classes. Formally, we train a conditional generator \( \mathcal{G}_\phi: \mathcal{Z} \times \mathbb{R}^K \rightarrow \mathcal{X} \), parameterized by \(\phi\), to invert the classifier’s behavior by optimizing it to minimize a composite loss

\[
\mathcal{L}_{\text{Inv}} = 
\alpha \cdot \mathcal{L}_{\text{KL}} +
\beta \cdot \mathcal{L}_{\text{CE}} +
\gamma \cdot \mathcal{L}_{\text{Cosine}}
\]

where, \( \mathcal{L}_{\text{KL}} \) is the KL Divergence loss, \( \mathcal{L}_{\text{CE}} \) is the Cross Entropy loss, and \( \mathcal{L}_{\text{Cosine}} \) is the Cosine Similarity loss. The hyperparameters \( \alpha, \beta, \gamma \) control the contribution of each individual loss term defined as:
\[
\mathcal{L}_{\text{KL}} = \sum_{i} P(i) \log \frac{P(i)}{Q(i)}, \quad \mathcal{L}_{\text{CE}} = -\sum_{i} y_{i} \log(\hat{y}_{i}), \quad 
\mathcal{L}_{\text{Cosine}} = \frac{1}{N(N-1)} \sum_{i \neq j} \cos(\theta_{ij})
\]

where \( \mathcal{L}_{\text{KL}} \) represents the KL Divergence between the input distribution \( P \) and the output distribution \( Q \), \( y_{i} \) is the set encoded label, \( \hat{y}_{i} \) is the predicted label from the classifier, and \( \cos(\theta_{ij}) \) is the cosine similarity between the features of generated images \( i \) and \( j \) in a batch of \( N \).

Given the vastness of the input space, during early training stages, these reconstructions tend to be visually incoherent and do not resemble real data, reflecting the model’s incomplete or uncertain understanding of the class manifolds. These reconstructions are assigned to the garbage class and added to the training set for the subsequent epochs. In subsequent epochs the classifier is trained using a weighted cross-entropy loss to account for the class imbalance introduced by addition of garbage samples.

By iteratively repeating this cycle of training, inversion, and exclusion, the model gradually learns to refine the decision boundaries while pushing anomalous content into the garbage class. As the training progresses, inverted samples in Fig \ref{7} begin to look like training data, indicating that the classifier has effectively carved out the in-distribution manifold while isolating outliers into the garbage class.

During inference, this training procedure equips the classifier to identify and reject out-of-distribution (OOD) inputs by assigning them to the garbage class. Additionally, the softmax confidence scores corresponding to class predictions can be used to assess the model's uncertainty. Low softmax confidence on in-distribution predictions indicates ambiguous or uncertain inputs, while high confidence in the garbage class suggests a strong belief that the input is OOD. We quantify uncertainty using the softmax confidence values across all \(n+1\) output classes by capturing how sharply peaked or spread out the model's predictive distribution is. The uncertainty estimate for a prediction \(\mathbf{p}\) is given by:
\begin{equation}
\text{UE}(\mathbf{p})=1 - \frac{\sum_{i=1}^{n+1} \left(p_i - \frac{1}{n+1}\right)^2}{\sum_{i=1}^{n+1} \left(\delta_{i,k} - \frac{1}{n+1}\right)^2}
\end{equation}
where \(k = \arg\max_i p_i\) and \(\delta_{i,k}\) is the Kronecker delta. The resulting score ranges from 0 to 1, providing an interpretable measure of confidence by computing the squared distance between the predicted vector \(\mathbf{p}\) and the uniform distribution, normalized by the maximum possible distance under a one-hot prediction.

\section{Quantitative Results}
We evaluate the effectiveness of our approach to uncertainty-aware out-of-distribution detection across four benchmark image classification datasets: MNIST \citep{deng2012mnist}, FashionMNIST \citep{xiao2017fashionmnistnovelimagedataset}, SVHN, and CIFAR-10 \citep{cf}. To assess OOD detection performance, we follow a one-vs-rest evaluation strategy: the model is trained exclusively on one dataset and evaluated on the remaining three as OOD sources.

\begin{table}[h]
\centering
\caption{Accuracy for both in and out-of-distribution datasets.}
\label{tab:results}{%
\begin{tabular}{lcccc}
\toprule
\textbf{Train \textbackslash\ Test} & \textbf{MNIST} & \textbf{FMNIST} & \textbf{SVHN} & \textbf{CIFAR-10} \\
\midrule
MNIST     & 99.1 & 89.5 & 99.1 & 99.4 \\
FMNIST    & 85.2 & 92.6 & 96.3 & 95.7 \\
SVHN      & 93.6 & 94.9 & 89.4 & 87.6 \\
CIFAR-10  & 97.8 & 95.7 & 88.2 & 85.5 \\
\bottomrule
\end{tabular}
}
\end{table}
Table~\ref{tab:results} presents the accuracy for uncertainty-aware OOD detection across all pairs of datasets. Each row corresponds to a model trained on one of the datasets and diagonal entries represent the in-distribution (ID) performance measured on the standard test set of the training dataset. Off-diagonal entries indicate OOD detection performance, where the accuracy represents how well the model distinguishes out-of-distribution samples by correctly classifying them into the garbage class. High values across both diagonal and off-diagonal entries demonstrate that the model maintains strong classification performance on ID data while reliably identifying OOD inputs. 

We also observe that while the majority of OOD samples are correctly assigned to the garbage class, a small percentage of the samples can still be misclassified into in-distribution classes. However, a significant finding is that the least confidently classified in-distribution sample is still more confidently classified compared to the most confidently misclassified out-of-distribution sample, suggesting the existence of a clear threshold. 

\section{Conclusion}
In conclusion, our unified framework seamlessly integrates out-of-distribution (OOD) detection and uncertainty estimation (UE) by extending the classification model with a garbage class and leveraging network inversion for inverted sample generation. Through iterative training and inversion cycles, the model learns to delineate in-distribution data from anomalous inputs while progressively refining its class boundaries. This approach enables robust OOD rejection and provides interpretable uncertainty estimates based on softmax confidence distributions. Future work can also consider the use n garbage classes—one for each of the in-distribution classes—for fine-grained separation of OOD samples and weighted individual OOD sample contribution to the loss while retraining the classifier based on uncertainty.

\bibliography{ref}
\bibliographystyle{plainnat}

\end{document}